\documentclass[letterpaper, 10 pt, conference]{ieeeconf}  

\IEEEoverridecommandlockouts                              

\usepackage{amsmath} 
\usepackage{amssymb}  
\usepackage{subfigure}
\usepackage{changes}
\usepackage{ulem}
\def\etal{{et al}.}

\title{\LARGE \bf
SegGrasp: Zero-Shot Task-Oriented Grasping via Semantic and Geometric Guided Segmentation
}

\author{Haosheng Li$^{*1}$ Weixin Mao$^{*2}$ Weipeng Deng$^{3}$ Chenyu Meng$^{1}$  Rui Zhang$^{4}$ \\ 
Fan Jia$^{5}$ Tiancai Wang$^{5}$  Haoqiang Fan$^{5}$ Hongan Wang$^{1}$ Xiaoming Deng$^{\dag1}$
\thanks{$^{*}$Equal contribution.}
\thanks{$^{\dag}$Corresponding author.}
\thanks{$^{1}$ Institute of Software, Chinese Academy of Sciences, Beijing, China
        }%
\thanks{$^{2}$ Waseda University, Tokyo, Japan
        }
\thanks{$^{3}$ University of Hong Kong, Hong Kong, China
        }
\thanks{$^{4}$ Zhejiang University, Zhejiang, China
        }
\thanks{$^{5}$ MEGVII Technology, Beijing, China
        }
}
\usepackage{algorithm}
\usepackage{algorithmic}
\usepackage{multirow}
\usepackage{graphicx}
\usepackage{booktabs}
\usepackage{colortbl}

\usepackage{pifont}
\newcommand{\cmark}{\ding{51}}%

\usepackage{xcolor}

\begin{document}

\maketitle
\thispagestyle{empty}
\pagestyle{empty}

\begin{abstract}

Task-oriented grasping, which involves grasping specific parts of objects based on their functions, is crucial for developing advanced robotic systems capable of performing complex tasks in dynamic environments. In this paper, we propose a training-free framework that incorporates both semantic and geometric priors for zero-shot task-oriented grasp generation. The proposed framework, SegGrasp, first leverages the vision-language models like GLIP for coarse segmentation. It then uses detailed geometric information from convex decomposition to improve segmentation quality through a fusion policy named GeoFusion. An effective grasp pose can be generated by a grasping network with improved segmentation. We conducted the experiments on both segmentation benchmark and real-world robot grasping. The experimental results show that SegGrasp surpasses the baseline by more than 15\% in grasp and segmentation performance.
\end{abstract}


\section{INTRODUCTION}

Recently, significant progress has been made in robotic grasping, resulting in the emergence of various grasping tasks, including task-oriented grasping~\cite{Learning-affordance,affordance-with-foundationmodels,Grasp-as-You-Say,shapegrasp,dexgraspnet,force-closure,grasp-d}. 
Task-oriented grasping selects the best area of an object to ensure successful and safe execution, especially for various tasks. For example, a robot should grasp a hammer with the head to hand over and the head to strike nails. Such functional distinctions are crucial for designing systems capable of executing complex context-aware grasping in many real-world applications.

In order to achieve task-oriented grasping, some traditional methods~\cite{affordance-with-foundationmodels,Semgrasp} learn prior knowledge of object grasping from collected data and generate grasp poses for different tasks based on these priors. Other knowledge-driven methods~\cite{shapegrasp, lan-grasp,tang2023graspgpt} take advantage of strong generalization abilities of large language models (LLMs) to perform reasoning and generate reasonable task-oriented grasps.  In addition, geometric information is also critical in task-oriented grasping. ShapeGrasp \cite{shapegrasp} uses convexity information of objects to determine the suitable part of the object to grasp. However, the heavy reliance on empirical heuristic algorithms may result in complex processes and inconsistent performance in varied scenarios. Insufficient utilization of geometric features as in~\cite{partslip,Constrained-6-DoF-Grasp,satr} also compromises the precision needed for task-oriented grasping. Furthermore, poor generalization to unseen objects also severely restricts the model's adaptability, making it challenging to perform task-oriented grasps in diverse environments. 


\begin{figure}[t]
    \centering
 \includegraphics[width=0.97\linewidth]{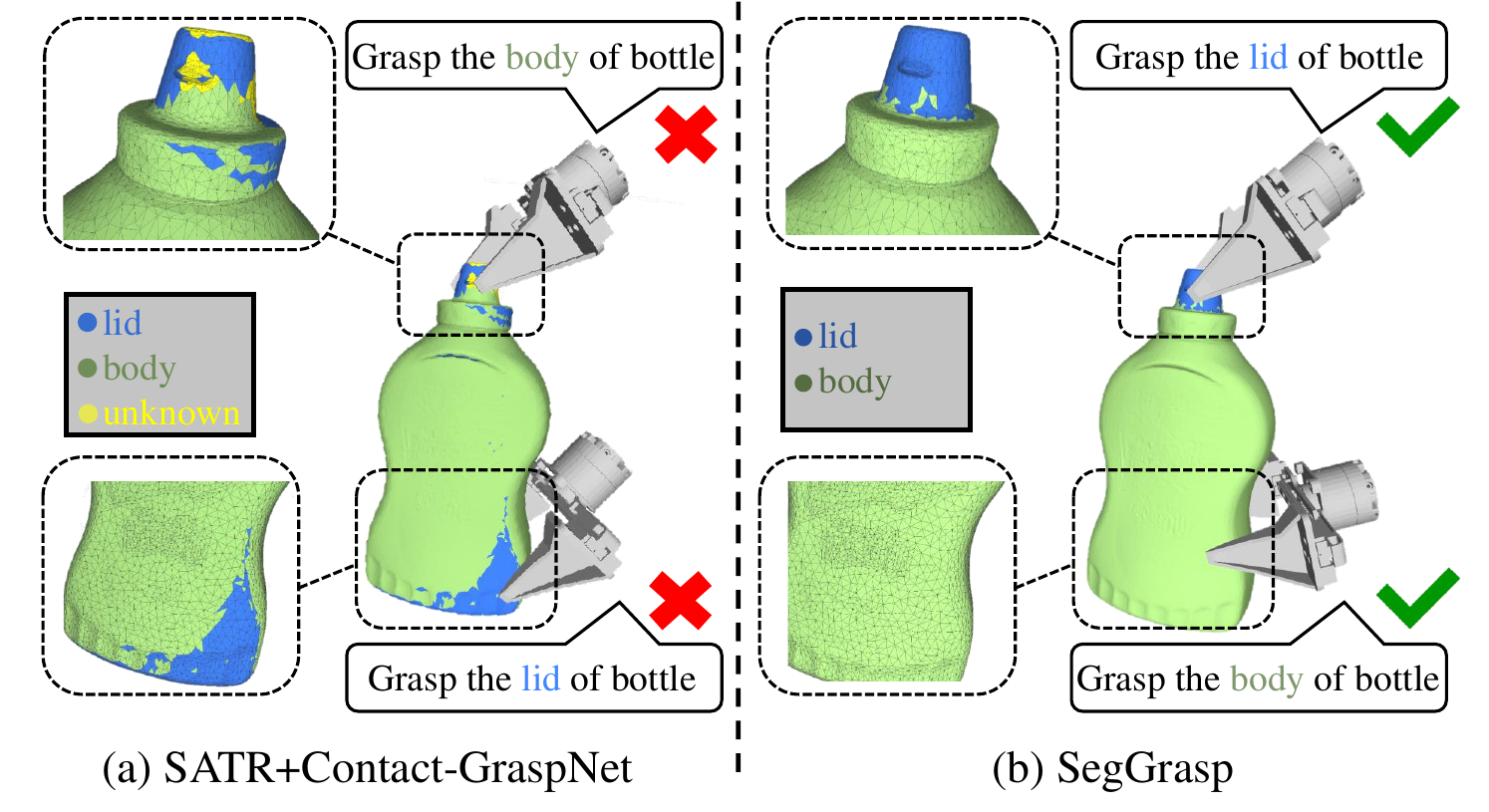}    
    \caption{\textbf{Comparison of the baseline method and our method.} The baseline method which consists of SATR and Contact-GraspNet often generates incorrect grasp poses due to mis-segmentation. In contrast, our approach produces cleaner and more precise segmentation without any unknown region, resulting in better grasp pose generation.}
    \label{fig_vis_grasp_seg}
\end{figure}
In this paper, to tackle with above limitation, we propose a novel training-free task-oriented grasping framework named SegGrasp. It leverages the robust generalization capabilities of zero-shot vision-language model to extract semantic priors of the input shape and incorporates detailed geometric priors through mesh convexity decomposition.
Specifically, our method begins with initial object coarse segmentation by employing open-vocabulary models such as Grounding DINO~\cite{groundingdino} and GLIP~\cite{glip}. Then we design a geometry-guided segmentation fusion policy named GeoFusion to refine the initial segmentation. The policy combines a multi-fusion process across various convex decompositions with a fine-grained optimization using the most detailed geometric information. This approach ensures precise and clear segmentation, which is crucial for accurate grasp generation. As shown in Fig.~\ref{fig_vis_grasp_seg}, our approach outperforms the baseline method, which combines the efficient zero-shot segmentation method SATR~\cite{satr} with a grasping network, in both segmentation and grasping quality. For example, our method produces a better segmentation of the bottle without any unknown regions as shown in Fig.~\ref{fig_vis_grasp_seg}(a) and more precise grasp. The final results of the segmentation can facilitate the existing grasping network such as Contact-GraspNet~\cite{contact-graspnet} to generate grasp poses with a high success rate. 

Our method demonstrates a strong zero-shot generalization capability, adaptable to new scenarios without any training. Using geometric information, SegGrasp achieves superior performance in task-oriented grasp generation tasks compared to image-based approaches~\cite{Constrained-6-DoF-Grasp,partslip,satr}. We conducted a comprehensive evaluation of the segmentation and grasping capabilities on ShapeNetPart \cite{shapenetpart} and our custom task-oriented grasping dataset. 
We exceeded our baseline method by a large margin in contact segmentation and grasping. 

Our contributions are summarized as follows. 
\begin{enumerate}
    \item We propose a novel zero-shot task-oriented grasping framework. This framework integrates segmentation and grasp generation in a fully-automated manner, significantly enhancing both efficiency and generalization ability.
    \item We design a segmentation fusion policy by fusing geometric priors and coarse segmentation to produce a refined
    contact segmentation for robotic grasp generation.
    \item Our work bridges the gap between segmentation models and grasp generation, demonstrating the practical viability of applying segmentation methods in the context of robotic grasping.
\end{enumerate}

\section{ RELATED WORKS}

\subsection{Task-oriented Grasp Generation}
Functional grasp generation aims to create grasps that facilitate specific tasks. Some methods rely on large datasets to learn grasp strategies.
OOAL~\cite{affordance-with-foundationmodels} learns the affordance of the object with the clip \cite{clip} based on the egocentric dataset~\cite{Object-based-affordances,Learning-affordance}. Some methods~\cite{Task-oriented-grasp,Grasp-as-You-Say,tang2023graspgpt,Semgrasp} , learn from annotated data to create functional grasp. FunctionalGrasp \cite{functionalgrasp} focuses on semantic hand-object interaction without detailed hand pose annotations. Wu \etal~\cite{Cross-Category-Functional-Grasp-Tansfer} introduces a cross-category grasp transfer method using object similarities and knowledge graphs for a new grasp synthesis. Although these data-driven methods perform well on existing data, their generalization to unseen scenarios is often limited. Moreover, Some methods focus on examining the nature and shape of objects to deduce their various functional parts. For example, LeRF-TOGO \cite{lerftogo} employs LeRF \cite{lerf} to construct a language field for three-dimensional scenes, associating each part of an object with its semantic meaning. ShapeGrasp \cite{shapegrasp} and Lan-Grasp \cite{lan-grasp} utilize a pretrained Large Language Model (LLM) \cite{gpt4} to infer parts of an object. Realdex \cite{realdex} uses Gemini as an evaluator to filter out grasps that do not meet specific functional requirements. Although these methods leverage sophisticated analytical techniques, their dependence on previous parameter settings \cite{shapegrasp} and extensive training times \cite{lerftogo} reduce their practical use.

\subsection{Zero-shot 3D Open-vocabulary Segmentation}
With the development of open vocabulary recognition methods~\cite{glip,groundingdino,groundingdino1.5}, zero-shot segmentation methods have advanced rapidly~\cite{satr,partslip,partslip++}. These approaches typically render point clouds or meshes in 2D images and utilize GLIP \cite{glip} for segmentation. Methods like \cite{pointclip,pointclipv2} utilize CLIP's \cite{clip} language sensitivity by querying the visual features of rendered images with linguistic features to generate segmentation results. However, these methods focus solely on extracting information from rendered images, neglecting the inherent geometric information in meshes or point clouds. PartDistill \cite{partdistill} incorporates a point cloud distillation component, significantly improving segmentation performance. ACD \cite{ACD} employs Approximate Convex Decomposition to generate self-supervised signals. However, both methods require pre-training on additional datasets. Therefore, there is still a lack of zero-shot segmentation methods that can produce accurate, refined, and unfragmented segmentation results, which are essential to generate functional grasps.

\subsection{Robotic Grasping using Geometric Information}
To date, geometric information is primarily used to generate physically stable grasps. The methods~\cite{force-closure,dexgraspnet} leverage geometric features of the mesh to compute force closure and generate stable grasps, while Pointclouds-Grasp-Detection~\cite{pointclouds-grasp-detection} also utilizes 3D geometric information in grasp generation. However, the use of geometric information in functional grasp generation is barely addressed. ShapeGrasp \cite{shapegrasp} employs convex decomposition as an initial object segmentation method. However, ShapeGrasp~\cite{shapegrasp} is mainly dependent on the quality of the initial convex decomposition. Unlike ShapeGrasp \cite{shapegrasp}, our work uses vision-language model~\cite{groundingdino,glip} as the initial segmentation method, utilizing geometric information from convex decomposition \cite{coacd} as further refinement.

\begin{figure*}[ht]
    \centering
 \includegraphics[width=\linewidth]{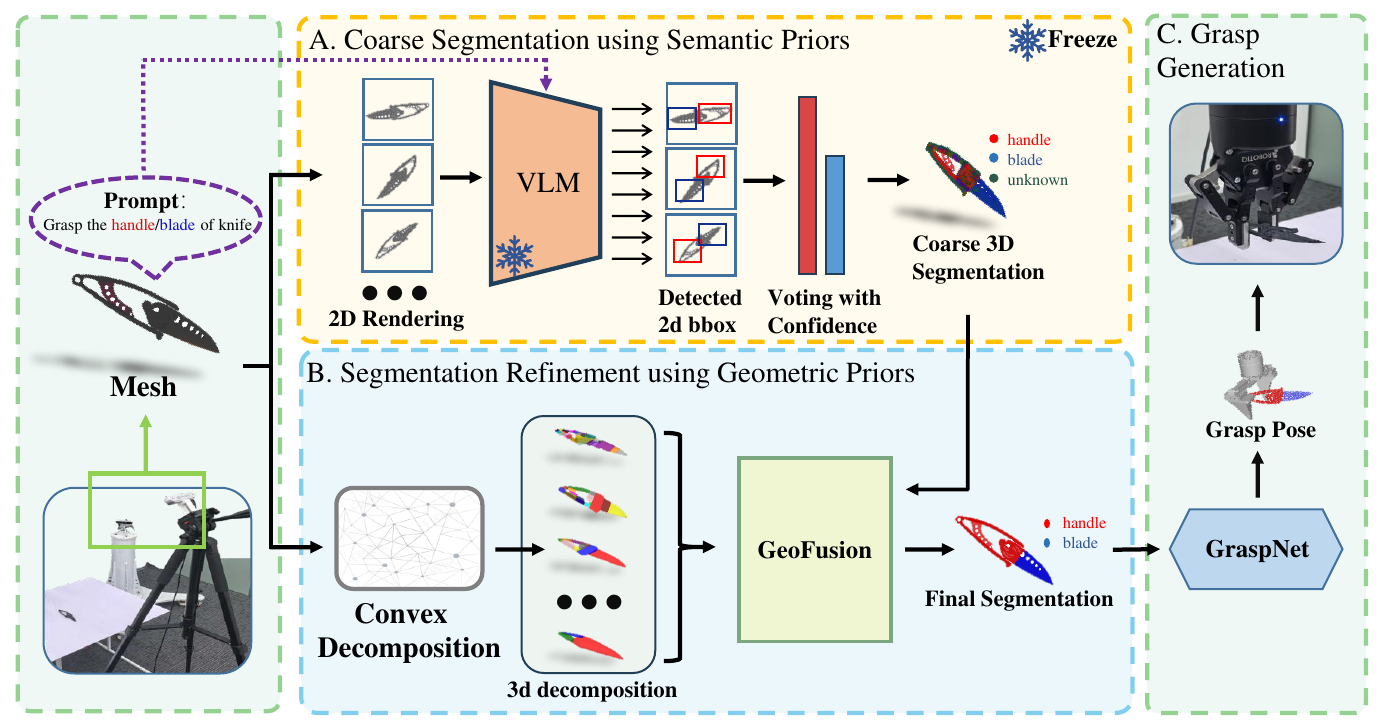}    
    \caption{\textbf{The overall architecture of SegGrasp.} Given a target object, our method renders the mesh from random viewpoints. We then utilize a vision-language model, such as Grounding DINO, to detect bounding boxes in the image and create coarse segmentation. The mesh is decomposed into multiple parts using various decomposition thresholds, each resulting in different segmentations. Using GeoFusion, these segmentations are fused with the initial coarse segmentation to achieve a refined segmentation. Finally, the refined segmentation faciliates Contact-GraspNet to generate high-quality grasp poses.}
    \label{fig_pipline}
\end{figure*}

\section{Method}


In this paper, we propose a novel zero-shot task-oriented grasp generation framework (see Fig.~\ref{fig_pipline}) that alleviates the limitations of existing methods by leveraging semantic priors via coarse segmentation and geometric priors through convex decomposition.
First, given the mesh of the object using a depth camera, we employ a vision-language model \cite{groundingdino,glip} to perform coarse segmentation of the object. Second, we use approximate convex decomposition (CoACD) \cite{coacd} to decompose the input shape into multiple convex parts, and design a segmentation refinement policy by fusing geometric priors named GeoFusion to produce a refined contact segmentation for robotic grasp generation. Finally, the corresponding grasp poses are generated by the Contact-GraspNet \cite{contact-graspnet} using the refined contact mask.


\subsection{Coarse Segmentation using Semantic Priors}


Using several calibrated Intel RealSense cameras, we obtain object point clouds from the depth image \cite{ICP,voxelnet} and reconstruct a 3D mesh through surface reconstruction techniques \cite{Poisson-surface-reconstruction}, which can be facilitated by various alternative approaches. Next, we efficiently obtain a coarse segmentation of the mesh by rendering it from several virtual viewpoints. Vision-language models such as \cite{groundingdino,glip} are then applied to generate 2D segmentation predictions for these rendered images. Finally, these 2D predictions are fused through confidence-based voting on the bounding boxes and projected back onto the mesh to produce the final segmentation result.

Given a mesh $M$ with $f$ faces and $m$ text prompts as input, we first render $v$ images of the input mesh from different viewpoints and feed text prompts into the vision-language model to generate bounding boxes in 2D images. We then map these prompt-specific bounding boxes back to the 3D mesh by identifying which faces fall into them from different viewpoints \cite{satr}. For each face, we compute a relevance score by summing its confidence scores across all viewpoints, which can be seen as a voting process. This process produces a score matrix, $\mathbf{S} = {S_{i,j}} \in \mathbb{R}^{f \times m}$, which evaluates the relevance of each face of the input mesh to the prompts and can be formulated as follows.
\begin{equation}
{S}_{i,j} =\sum_{m}{V(i,bbox_{m,j}) \times \text{confidence}_{m,j}}  \label{eq1}
\end{equation}

\noindent where $bbox_{m,j}$ and $\text{confidence}_{m,j}$ represent the bounding box and confidence score for the $j$-th text prompt in the $m$-th rendered viewpoint respectively, both predicted by the VLM model \cite{groundingdino}\cite{glip}, and $V(i, bbox)$ denotes the pixel count within the $i$-th face of input mesh inside bounding box $bbox$.

The coarse segmentation intuitively ensures that the larger area of rendered faces with a high confidence score contributes more to the segmentation prediction, thus maintaining the overall accuracy of the rough segmentation. 
 \begin{figure}[th]
    \centering
 \includegraphics[width=\linewidth]{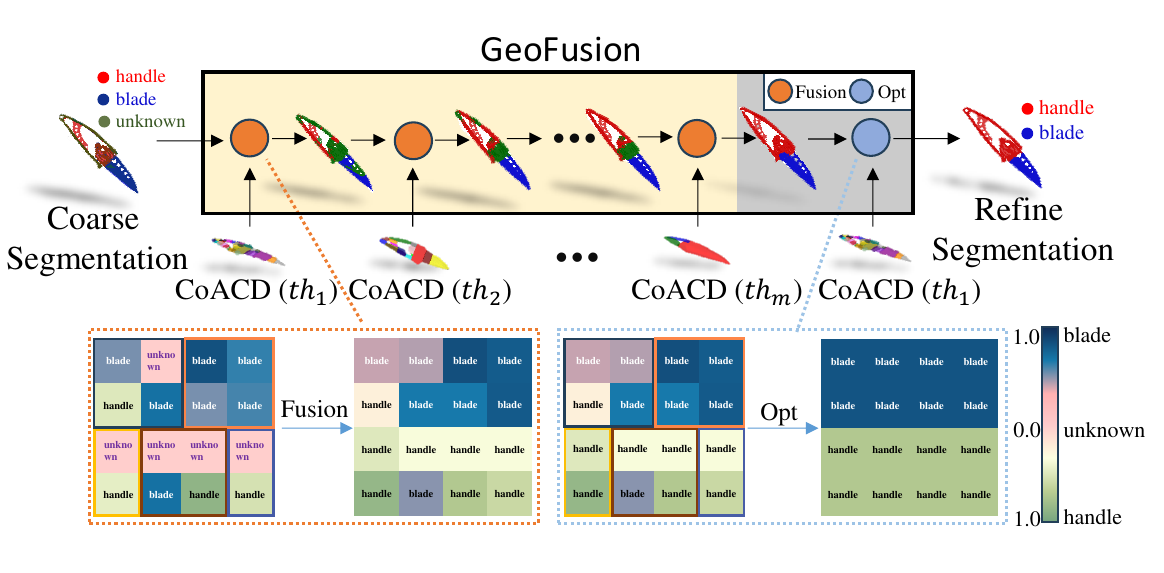}    
 
    \caption{\textbf{The detail of GeoFusion.} The upper section shows GeoFusion starting with coarse segmentation and convex decomposition under different decomposition thresholds. After multi-fusion ('Fusion') and fine-grained optimization ('Opt'), the unknown parts of the knife decrease, and the segmentation results improves. 
    The color of each square indicates the category score of the faces, corresponding to the value in the matrix $S$. Fine-grained optimization can effectively enforce that faces within the same segment belong to the same object part.}

    \label{method_explain_new}
\end{figure}

\subsection{Segmentation Refinement using Geometric Priors}
Although coarse segmentation is effective for vision-level semantic segmentation, occlusion and misrecognition will introduce substantial segmentation errors and many unknown regions, which is unacceptable for grasp generation (see Fig.~\ref{fig_vis_grasp_seg}(a)). To address these issues, we use the intrinsic shape geometry of the input mesh for segmentation refinement.

Generally, the functional grasp regions of an object typically have a strong correlation with its intrinsic geometric properties~\cite{shapegrasp,ACD,lerftogo}. Convex decomposition can split an object into parts with similar geometric properties \cite{coacd}. Therefore, faces belonging to the same convex decomposition part are more likely to correspond to the same functional grasping region. Following this hypothesis~\cite{shapegrasp,coacd}, we propose the GeoFusion policy (see Fig.~\ref{method_explain_new}). By aggregating across different decomposition thresholds, GeoFusion fuses geometric priors with the initial segmentation scores. It adjusts the scores to enhance consistency within each part, then applies the finest convex decomposition for final supervision, producing the final segmentation result.
\subsubsection{Shape Convex Decomposition}
We first use CoACD~\cite{coacd} to divide the input mesh $M$ into $n$ convex parts $\{{P}_i\}_{i=1}^n$, that is. 
\begin{equation}
M =\bigcup_{k=1}^{n} {P(k,th)}
\label{eq3}
\end{equation}
where  $P(k,th)$ is the $k$-th part of the convex decomposition result under decomposition threshold $th$. As shown in Fig.~\ref{method_explain_new}, the bounding box represents the convex decomposition result where the square corresponds to the face of the input mesh.

\subsubsection{Geometry-Guided Segmentation Fusion Policy}
Given the coarse segmentation scores and shape convex decomposition under various decomposition threshold $[th_1,\cdots,th_m]$, we design a segmentation refinement policy named GeoFusion using geometric and semantic clues, consisting multi-fusion (orange part in Fig.~\ref{method_explain_new}) and fine-grained optimization (grey part in Fig.~\ref{method_explain_new}).

For the $k$-th part of decomposed regions under the same decomposition threshold $th_w$, 
we calculate the overall area-weighted relevance score $S_{rev}(k,th_w)$ with respect to the query prompt. 
\begin{equation}
S_{rev}(k,th_w) = \frac{\sum_{u \in P(k,th_w) }{ area(f_u)\times\text{S}_{u,*}}}{l \times \sum_{u \in P(k,th_w)}{area(f_u)}}
\label{eq3}
\end{equation}

\noindent  where $area(f_u)$ indicates the area of the $u$-th face of the input mesh, $l$ is the number of faces in the $k$-th part of decomposition.

Then, the relevant scores $\hat{S}_{i,*}$ of the $i$-th faces within this convex decomposition are adjusted by adding the overall relevant score. We design a multi-fusion process across various convex decomposition thresholds, the faces within the same part are more likely to belong to the same category.
\begin{equation*}
\hat{S}_{i,*}=S_{i,*} + \sum_{w=1}^{m} \sum_{k} S_{rev}(k,th_w)\times \mathbb{I}(i \in P(k,th_w))
\end{equation*}
where $\mathbb{I}$ is an indicator function defined as:
\begin{equation}
\mathbb{I}(i \in P(k,th_w)) =
\begin{cases}
1 & \text{if } i\text{-th face of mesh} \in P(k,th_w) ,\\
0 & \text{otherwise}.
\end{cases}
\label{eq4}
\end{equation}

After applying this multi-fusion across different convex decomposition thresholds, we achieve more precise segmentation results (see Section \ref{subsec:Ablation}).

Despite the significant segmentation improvements provided by above multi-fusion process, there are still occasional instances of incorrect grasp region identification in real-world grasping experiments. These errors are primarily due to occlusions and segmentation inaccuracies during rendering, leading to mixed segmentation results. Although such issues may be minor in the context of segmentation, they can cause critical errors in functional grasp generation.

To address this issue, we use a fine-grained optimization method to refine segmentation, which uses the results of detailed convex decomposition with the smallest convexity threshold $th_1$. We calculate the overall score $\hat{S}_{rev}(k,th_1)$ based on the refined segmentation score matrix $\hat{S}$ as follows.
\begin{equation}
\hat{S}_{rev}(k,th_1) = \sum_{u \in P(k,th_{1})}{ area(f_u)\times\hat{S}_{u,*}}
\label{eq7}
\end{equation}

Next, we ensure that the score for each face within the same part of the finest convex decomposition is consistent with the overall score (see 'Opt' process in Fig.~\ref{method_explain_new}):
\begin{equation}
\hat{S}_{i,*}=\sum_{k}{\hat{S}_{rev}(k,th_{1})\times\mathbb{I}(i \in P(k,th_{1}))}
\label{eq8}
\end{equation}

The final result of the $i$-th faces of the input mesh is determined by:
\begin{equation}
{R}_{i} = \text{arg max}(\hat{S}_{i,*})
\label{eq9}
\end{equation}
where ${R}_{i}$ is the final result of the $i$-th faces of the input mesh, $P(k,th_{1})$ represents the $k$-th part resulting from the finest convex decomposition. According to Eq.~\ref{eq8} and Eq.~\ref{eq9}, the faces within each decomposition part are classified as the same functional grasping part, improving the performance in grasping experiments (see Section \ref{subsec:Comparison with State-of-the-art} and \ref{subsec:Ablation}).

\subsection{Grasp Generation}
Once the final segmentation results are obtained, we employ one of the state-of-the-art grasp generation methods--Contact-GraspNet \cite{contact-graspnet} to generate grasp poses. Since Contact-GraspNet uses point clouds as input, we uniformly sample point clouds from each input mesh to facilitate grasp generation. With the grasp poses generated, we use our segmentation results to select the proper functional grasp poses corresponding to each input prompt.
\begin{table*}[t]
\caption{Comparison of partial segmentation results on the ShapeNetPart dataset using different methods.}
\vspace{-0.3cm}
\setlength{\tabcolsep}{5pt}
\label{table1}
\begin{center}
\begin{tabular}{c|c|c|cccccccccccc}
\specialrule{1pt}{0pt}{1pt}
\toprule
Method & Backbone & mIoU(\%) & knife & lamp & airplane & mug & guitar & pistol & bag& earphone & cap & chair& table \\
\hline
3DH~\cite{3dh} & CLIP~\cite{clip} & 5.84 & 1.58 & 13.21 & 5.81 & 0.65 & 0.86 & 1.36 & 2.05 & 9.55 & 2.85 &15.53 &10.77 \\ 
SATR~\cite{satr} & GLIP~\cite{glip}& 34.16 & 45.92 & 30.22 & 38.46 & 52.31 & 40.22 & 20.87 & 44.56 & 16.9 & 24.01 & 33.16 & 31.41 \\
\hline
\rowcolor[gray]{.9} Ours &GLIP~\cite{glip}& 49.01 &62.47 &52.39 &\textbf{56.09} &79.22 &69.18 &\textbf{25.54} & \textbf{56.62} &19.39 &26.36 & \textbf{53.87} &37.97 \\
\rowcolor[gray]{.9} Ours &Grounding DINO~\cite{groundingdino}& \textbf{53.14} & \textbf{69.11} & \textbf{56.06} &53.27 &\textbf{83.66} & \textbf{75.57} & 24.04 &55.00 & \textbf{29.03} & \textbf{43.09} &50.96 & \textbf{44.80}\\
\bottomrule
\specialrule{1pt}{1pt}{0pt}
\end{tabular}
\end{center}
\end{table*}

\section{EXPERIMENT}

\subsection{Dataset and Metric}

We conducted comparative analyzes of our method against existing approaches on two datasets: the segmentation dataset ShapeNetPart \cite{shapenetpart} and our custom-built grasping dataset SegGraspSet.

\vspace{1mm}
\noindent \textbf{ShapeNetPart} \cite{shapenetpart} is a pre-annotated segmentation benchmark containing approximately 800 objects annotated from ShapeNet \cite{shapenet}. This dataset serves as a robust benchmark for assessing segmentation performance, providing a diverse range of shapes and configurations that test the generalization ability and accuracy of segmentation algorithms. 

\begin{figure}[h]
    \centering
 \includegraphics[width=\linewidth]{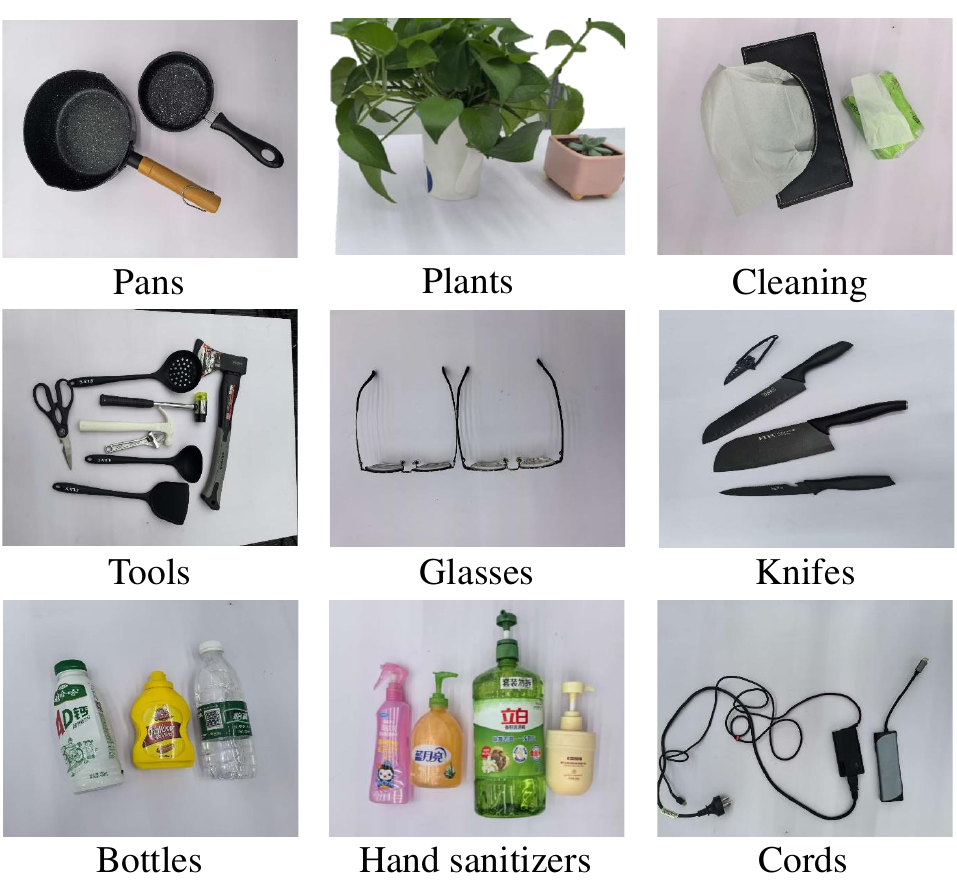}
    
    \caption{Overview of our SegGraspNet dataset which includes 9 common categories from everyday life.}
    \label{seggraspnet}
\end{figure}

\vspace{1mm}
\noindent \textbf{SegGraspSet} is our custom-built functional grasping dataset, inspired by LERF-TOGO~\cite{lerftogo} and ShapeGrasp~\cite{shapegrasp}, which comprises 9 categories with a total of 29 objects as shown in Fig.~\ref{seggraspnet}, each featuring distinct functional parts. This dataset is specifically designed and collected to evaluate the ability of grasping methods to understand and generate grasps that are not just geometrically feasible but also functionally appropriate.

\vspace{1mm}
\noindent \textbf{Evaluation Metrics.} For ShapeNetPart \cite{shapenetpart}, we use the mean Intersection over Union (mIoU) metric to evaluate segmentation accuracy. For SegGraspSet, we use part selection accuracy (Part Sel.) as the primary metric. Additionally, we assess the grasp success rate by selecting the top-10 grasp poses based on confidence rankings and lifting the object to a height of 20 cm, as shown in Fig.~\ref{fig_overview}. Since we conducted validation on two robotic systems, the overall grasp success rate (Suc.) is calculated averagely from both systems. Incorrectly selecting the wrong parts and failing to grasp the correct ones are both considered failures. In addition, we calculate the variance of the generated grasp poses to evaluate the diversity of grasp configurations of the robotic hand. In our application, we convert the generated grasp poses to quaternions and calculate their variance. Higher variance reflects more diverse grasp poses generated.

\subsection{Implementation Details}
\begin{figure}[h]
    \centering
    \subfigure[Robot experiment setup]{
        \includegraphics[width=\linewidth]{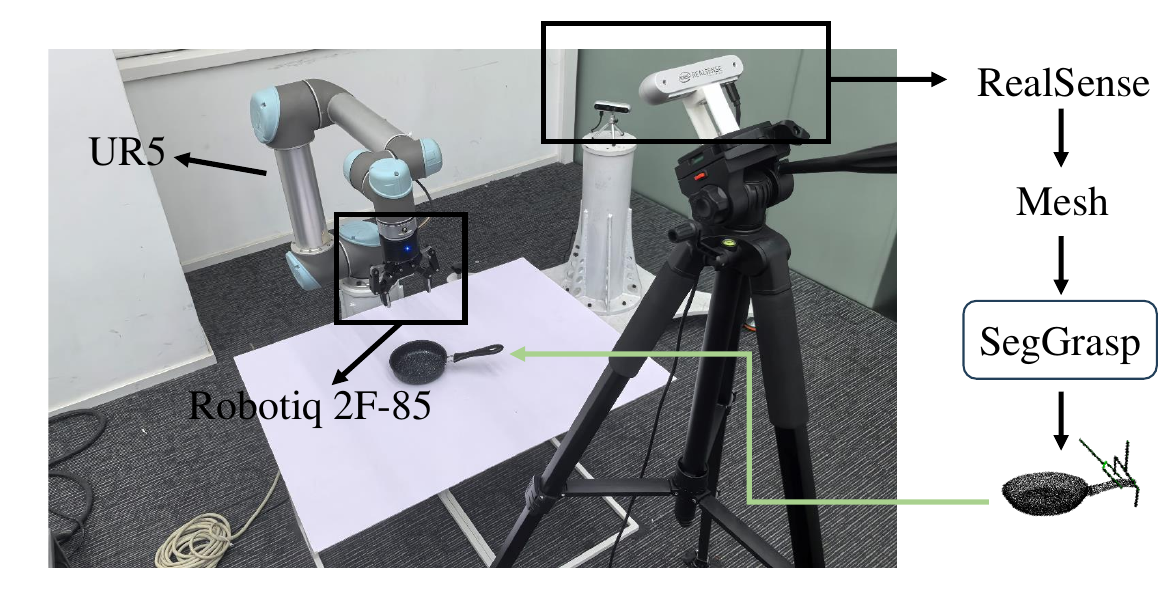}
        
        \label{fig_experiment}
        
    }


    \subfigure[Illustration of grasp results ]{
        \includegraphics[width=\linewidth]{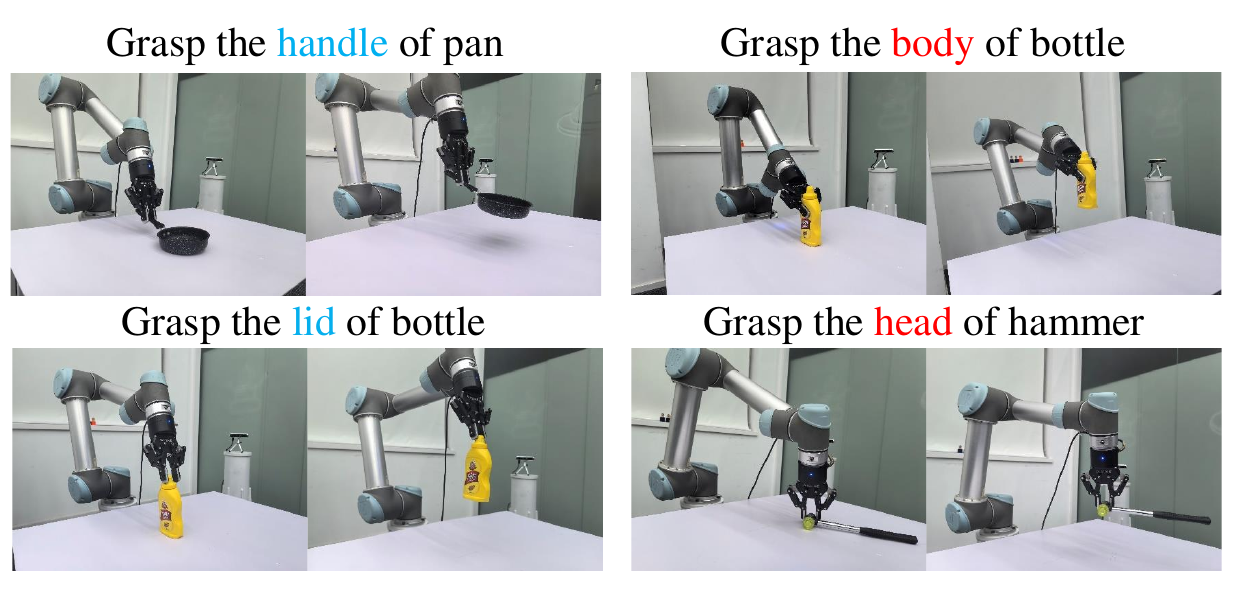}
        \label{fig_overview}
    }

    \caption{\textbf{The details of our robot experiments.} (a) shows one of our robot grasping setups, consisting of a UR5 arm, RealSense D456 camera, and Robotiq gripper. (b) demonstrates grasp results from our method on the SegGraspSet dataset.}
\end{figure}

During the coarse segmentation stage, we render images from 10 distinct viewpoints and employ GLIP \cite{glip} and Grounding DINO~\cite{groundingdino} to generate bounding boxes. For convex decomposition, we use the CoACD~\cite{coacd} method, selecting convexity thresholds from 0.01 to 0.25 with step 0.01. If the segmentation generated by the adjacent convexity thresholds is identical, we use the result from the previous threshold until we obtain distinct segmentation. In the grasp generation phase, 
we uniformly sample 30, 000 points on each mesh and employ Contact-GraspNet \cite{contact-graspnet} to generate grasps. The object part that contains the closest surface face to the generated grasp point is considered as the functional grasping part. We evaluate our framework on two robotic systems. The first system employs a UR5 robotic arm with 6 joints and a Robotiq gripper, as shown in Fig.~\ref{fig_experiment}. The second system employs an ARX5 robotic arm with a parallel gripper which is shown in our demonstration video. Since the grasp pose generation method of ShapeGrasp \cite{shapegrasp} directly targets the object's centroid and uses a fixed gripper rotation that is perpendicular to the table during grasping, the success rate is defined as the percentage of successful grasps over ten attempts. And the variance of the grasp pose is ignored to ensure a fair comparison with our methodology.

\subsection{Comparison with State-of-the-art}
\label{subsec:Comparison with State-of-the-art}
\subsubsection{Segmentation Results}
In Table~\ref{table1}, we compare our segmentation results with SATR~\cite{satr} and 3DH~\cite{3dh}. We observe that our approach outperforms the baseline by a large margin (50\%) on the ShapeNetPart dataset \cite{shapenetpart}. In addition, Grounding DINO \cite{groundingdino} serves as a better backbone for segmentation in most categories of test objects.
\subsubsection{Grasp Results}
In a real-world experiments as shown in Fig.~\ref{fig_overview}, we conducted a comparative analysis of our method against state-of-the-art zero-shot functional grasping techniques ShapeGrasp \cite{shapegrasp} and baseline method which is Contact-GraspNet~\cite{contact-graspnet} with segmentation by SATR \cite{satr}.


\begin{table}[h]
\caption{Comparison of the grasp results using different methods. $*$-G uses GLIP as its VLM backbone, while  $*$-D uses Grounding DINO as its VLM backbone.}
\label{table2}
\begin{center}
\begin{tabular}{l|c|c|c}
\toprule
Method & Part Sel. $\uparrow$  & Suc. $\uparrow$ & $\sigma^2$ $\uparrow$ \\ 
\midrule

 Baseline-G &0.67 & 0.33 & 0.160 \\
 Baseline-D &0.71 & 0.37 & 0.169 \\
 ShapeGrasp \cite{shapegrasp} &0.80 & 0.39  & - \\
  SegGrasp-G & 0.83 & 0.45 & 0.180 \\
  SegGrasp-D & \textbf{0.87} & \textbf{0.46}  & \textbf{0.184} \\
\bottomrule
\end{tabular}
\end{center}
\end{table}

The comparative analysis in Table~\ref{table2} demonstrates the superior performance of our methods, where SegGrasp-G and SegGrasp-D use the Grounding DINO \cite{groundingdino} and GLIP \cite{glip} as the vision-language model, respectively. Our approach achieves higher part selection accuracy, grasp success rates, and grasp pose diversity. As shown in Fig.~\ref{fig_vis_grasp_seg}, SATR \cite{satr} which only relies on 2D images, struggles with occlusion and intermixing, leading to unknown regions and fewer grasp generation. These issues reduce the grasp stability and accuracy, which leads the lower performance of SATR \cite{satr}, aligning with our experimental results.

Additionally, we found that ShapeGrasp's efficacy is limited by its sensitivity to convex decomposition and its centroid-based grasping strategy. Moreover, Grounding DINO proves to be more effective than GLIP overall. These findings highlight the robustness and adaptability of SegGrasp, demonstrating its effectiveness for zero-shot functional grasping tasks in real-world scenarios.
More results can be found in the supplementary video.

\subsection{Ablation Study}
\label{subsec:Ablation}

We conducted an ablation study on SegGrasp to evaluate the impact of each component across different metrics including mIoU on ShapeNetPart \cite{shapenetpart} and the accuracy of part selection (Part Sel.) and success grasp rate (Suc.) on SegGraspSet. As shown in Table \ref{table4}, coarse segmentation (coarse) is the Baseline-D in Table \ref{table2}, while adding the multi-fusion process (orange part in Fig.~\ref{method_explain_new}) significantly improved performance. Fine-grained optimization (grey part in Fig.~\ref{method_explain_new}) has a small impact on mIoU, but crucially prevents errors in part identification and grasp generation. Combining both components leads to optimal performance, demonstrating their complementary roles and validating the effectiveness of our GeoFusion policy.
\begin{table}[h]
\setlength{\tabcolsep}{5pt}
\centering
\caption{Effect of SegGrasp components, multi-fusion and fine-Opt. Fine-Opt: fine-grained optimization. 
}
\begin{tabular}{c c c | c c c}
\toprule
Coarse & Multi-Fusion & Fine-Opt & mIoU $\uparrow$ & Part Sel. $\uparrow$ & Suc. $\uparrow$\\ 
\midrule
\cmark &  &  & 0.34 & 0.71 & 0.37\\ 
 \cmark& \cmark &  & 0.46 & 0.84 & 0.42\\ 
 \cmark&  & \cmark & 0.42 & 0.75 & 0.39\\ 
\cmark & \cmark & \cmark & \textbf{0.47 }& \textbf{0.87} & \textbf{0.46}\\ 
 \bottomrule
\end{tabular}
\label{table4}
\end{table}




In order to further evaluate the effect of GeoFusion, we design an alternative segmentation fusion strategy named GeoSpreading for comparison.    
GeoSpreading involves propagating the segmentation score from fine to coarse segmentation, derived from convex decomposition at different convexity thresholds. Unlike GeoFusion, GeoSpreading updates $S$ across different convexity thresholds. GeoSpreading updates the coarse segmentation scores as follows.
\begin{equation}
\hat{S}_{i,*}=\hat{S}_{i,}+\frac{\sum_{u \in P(k,th) }{ area(f_u)\times\hat{S}_{u,*}}}{m \times \sum_{u \in P(k,th)}{area(f_u)}},i \in P(k,th)
\label{eq5}
\end{equation}

\begin{figure}[h]
    \centering
 \includegraphics[width=0.95\linewidth]{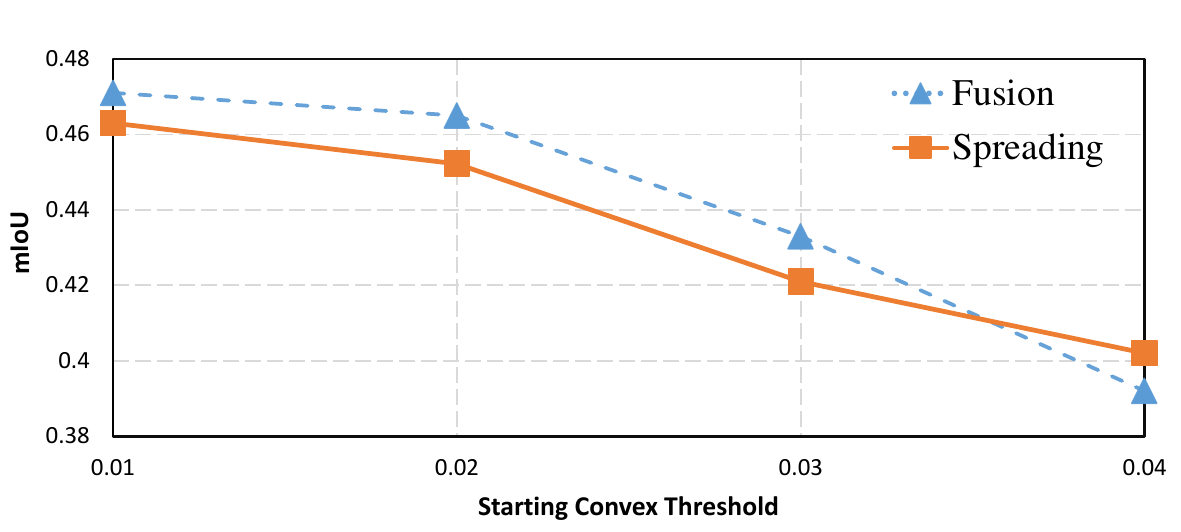}
    
    \caption{Performance comparison of GeoFusion ('Fusion') and GeoSpreading ('Spreading') techniques across varying convex thresholds, demonstrating their impact on mIoU.}
    \label{ablation}
\end{figure}

We evaluated these methods on ShapeNetPart \cite{shapenetpart}. As shown in Fig.~\ref{ablation}, GeoFusion consistently outperforms GeoSpreading at different decomposition thresholds. In addition, performance improves as the initial decomposition threshold decreases, indicating that effective segmentation is based heavily on fine-grained information.

\section{CONCLUSIONS}
We propose a zero-shot task-oriented grasping framework, which generates grasp poses for different object parts based on different task prompts. Our framework achieves outstanding performance across various tasks by integrating both semantic and geometric priors. Furthermore, it demonstrates strong generalization capabilities with the robust zero-shot vision-language model. Experimental results show that our approach outperforms the compared methods by a large margin. In future work, we plan to apply our framework to dexterous hands. Its ability to precisely segment contact areas provides a great advantage, especially given the larger surface interactions in dexterous grasping. We expect our approach to enhance performance in complex manipulation tasks, advancing robotic dexterity and autonomy.


\bibliographystyle{IEEEtran}









\bibliography{main}

\end{document}